\documentclass[10pt, a4paper]{article}
\usepackage{lrec}
\usepackage{multibib}
\newcites{languageresource}{Language Resources}
\usepackage{graphicx}
\usepackage{tabularx}
\usepackage{soul}
\usepackage{amsmath}
\usepackage{multirow}
\usepackage{booktabs}
\usepackage{url}
\usepackage{array}
\usepackage{epstopdf}
\usepackage[latin1]{inputenc}
\usepackage{hyperref}
\usepackage{xstring}

\title{Dataset for the First Evaluation on Chinese Machine Reading Comprehension}
\name{Yiming Cui$^\dag$, Ting Liu$^\ddag$, Zhipeng Chen$^\dag$, Wentao Ma$^\dag$, Shijin Wang$^\dag$ and Guoping Hu$^\dag$}

\address{
$^\dag$Joint Laboratory of HIT and iFLYTEK, iFLYTEK Research, Beijing, China \\
{$^\ddag$Research Center for Social Computing and Information Retrieval,}\\
{Harbin Institute of Technology, Harbin, China}\\
{$^\dag$\tt\{ymcui,zpchen,wtma,sjwang3,gphu\}@iflytek.com}\\  
{$^\ddag$\tt tliu@ir.hit.edu.cn}\\
}

\abstract{
Machine Reading Comprehension (MRC) has become enormously popular recently and has attracted a lot of attention.
However, existing reading comprehension datasets are mostly in English.
To add diversity in reading comprehension datasets, in this paper we propose a new Chinese reading comprehension dataset for accelerating related research in the community. 
The proposed dataset contains two different types: cloze-style reading comprehension and user query reading comprehension, associated with large-scale training data as well as human-annotated validation and hidden test set.
Along with this dataset, we also hosted the first Evaluation on Chinese Machine Reading Comprehension (CMRC-2017) and successfully attracted tens of participants, which suggest the potential impact of this dataset.
 \\ \newline }

\begin{document}

\maketitleabstract

%%%%%%%%%%%%%%%%%%%%%%%
\section{Introduction}
Machine Reading Comprehension (MRC) has become enormously popular in recent research, which aims to teach the machine to comprehend human languages and answer the questions based on the reading materials. Among various reading comprehension tasks, the cloze-style reaing comprehension is relatively easy to follow due to its simplicity in definition, which requires the model to fill an exact word into the query to form a coherent sentence according to the document material.
Several cloze-style reading comprehension datasets are publicly available, such as CNN/Daily Mail \cite{hermann-etal-2015}, Children's Book Test \cite{hill-etal-2015}, People Daily and Children's Fairy Tale \cite{cui-etal-2016}. 

In this paper, we provide a new Chinese reading comprehension dataset\footnote{CMRC 2017 Public Datasets: {\url{https://github.com/ymcui/cmrc2017}}.}, which has the following features
\begin{itemize}
	\item We provide a large-scale automatically generated Chinese cloze-style reading comprehension dataset, which is gathered from children's reading material.
	\item Despite the automatic generation of training data, our evaluation datasets (validation and test) are annotated manually, which is different from previous works.
	\item To add more diversity and for further investigation on transfer learning, we also provide another evaluation datasets which is also annotated by human, but the query is more natural than the cloze type.
\end{itemize}

We also host the 1st Evaluation on Chinese Machine Reading Comprehension (CMRC2017), which has attracted over 30 participants and finally there were 17 participants submitted their evaluation systems for testing their reading comprehension models on our newly developed dataset, suggesting its potential impact. We hope the release of the dataset to the public will accelerate the progress of Chinese research community on machine reading comprehension field. 

We also provide four official baselines for the evaluations, including two traditional baselines and two neural baselines. In this paper, we adopt two widely used neural reading comprehension model: AS Reader \cite{kadlec-etal-2016} and AoA Reader \cite{cui-acl2017-aoa}.

The rest of the paper will be organized as follows. In Section 2, we will introduce the related works on the reading comprehension dataset, and then the proposed dataset as well as our competitions will be illustrated in Section 3. The baseline and participant system results will be given in Section 4 and we will made a brief conclusion at the end of this paper.

        \begin{table*}[ht]
        \begin{center}
        \begin{tabular}{lrrrrrr}
        \toprule
        & & \multicolumn{2}{c}{Cloze Track} & \multicolumn{2}{c}{User Query Track} \\
        & Train & Validation & Test & Validation & Test \\
        \midrule
        \# Query & 354,295 & 2,000 & 3,000 & 2,000 & 3,000 \\
        Max \# tokens in docs & 486 & 481 & 484 & 481 & 486 \\
        Max \# tokens in query & 184 & 72 & 106 & 21 & 29 \\
        Avg \# tokens in docs & 324 & 321 & 307 & 310 & 290 \\
        Avg \# tokens in query & 27 & 19 & 23 & 8 & 8 \\
        Vocabulary & \multicolumn{5}{c}{94,352}  \\
        \bottomrule
        \end{tabular}
        \end{center}
        \caption{\label{data-stats} Statistics of the dataset for the 1st Evaluation on Chinese Machine Reading Comprehension (CMRC-2017).}
        \end{table*}

%%%%%%%%%%%%%%%%%%%%%%%
\section{Related Works}\label{related-works}
In this section, we will introduce several public cloze-style reading comprehension dataset.

%%%%%%%
\subsection{CNN/Daily Mail}
Some news articles often come along with a short summary or brief introduction. Inspired by this, Hermann et al. \shortcite{hermann-etal-2015} release the first cloze-style reading comprehension dataset, called CNN/Daily Mail\footnote{The pre-processed CNN and Daily Mail datasets are available at {\url{http://cs.nyu.edu/~kcho/DMQA/}}}. Firstly, they obtained large-scale CNN and Daily Mail news data from online websites, including main body and its summary. Then they regard the main body of the news as the {\em Document}. The {\em Query} is generated by replacing a name entity word from the summary by a placeholder, and the replaced named entity word becomes the {\em Answer}. 
Along with the techniques illustrated above, after the initial data generation, they also propose to anonymize all named entity tokens in the data to avoid the model exploit world knowledge of specific entities, increasing the difficulties in this dataset.
However, as we have known that world knowledge is very important when we do reading comprehension in reality, which makes this dataset much artificial than real situation. Chen et al. \shortcite{chen-etal-2016} also showed that the proposed anonymization in CNN/Daily Mail dataset is less useful, and the current models \cite{kadlec-etal-2016,chen-etal-2016} are nearly reaching ceiling performance with the automatically generated dataset which contains much errors, such as coreference errors, ambiguous questions etc.

%%%%%%%
\subsection{Children's Book Test}
Another popular cloze-style reading comprehension dataset is the Children's Book Test (CBT)\footnote{Available at {\url{http://www.thespermwhale.com/jaseweston/babi/CBTest.tgz}}} proposed by Hill et al. \shortcite{hill-etal-2015} which was built from the children's book stories. Though the CBT dataset also use an automatic way for data generation, there are several differences to the CNN/Daily Mail dataset. They regard the first 20 consecutive sentences in a story as the {\em Document} and the following 21st sentence as the {\em Query} where one token is replaced by a placeholder to indicate the blank to fill in. Unlike the CNN/Daily Mail dataset, in CBT, the replaced word are chosen from various types: Name Entity (NE), Common Nouns (CN), Verbs (V) and Prepositions (P). The experimental results showed that, the verb and preposition answers are not sensitive to the changes of document, so the following works are mainly focusing on solving the NE and CN genres.

%%%%%%%
\subsection{People Daily \& Children's Fairy Tale}
The previously mentioned datasets are all in English. To add diversities to the reading comprehension datasets, Cui et al. \shortcite{cui-etal-2016} proposed the first Chinese cloze-style reading comprehension dataset: People Daily \& Children's Fairy Tale, including People Daily news datasets and Children's Fairy Tale datasets.
They also generate the data in an automatic manner, which is similar to the previous datasets. They choose short articles (several hundreds of words) as {\em Document} and remove a word from it, whose type is mostly named entities and common nouns. Then the sentence that contains the removed word will be regarded as {\em Query}. 
To add difficulties to the dataset, along with the automatically generated evaluation sets (validation/test), they also release a human-annotated evaluation set. The experimental results show that the human-annotated evaluation set is significantly harder than the automatically generated questions. The reason would be that the automatically generated data is accordance with the training data which is also automatically generated and they share many similar characteristics, which is not the case when it comes to human-annotated data.

%%%%%%%%%%%%%%%%%%%%%%%

\begin{figure*}[t]
  \centering
  \includegraphics[width=1\textwidth]{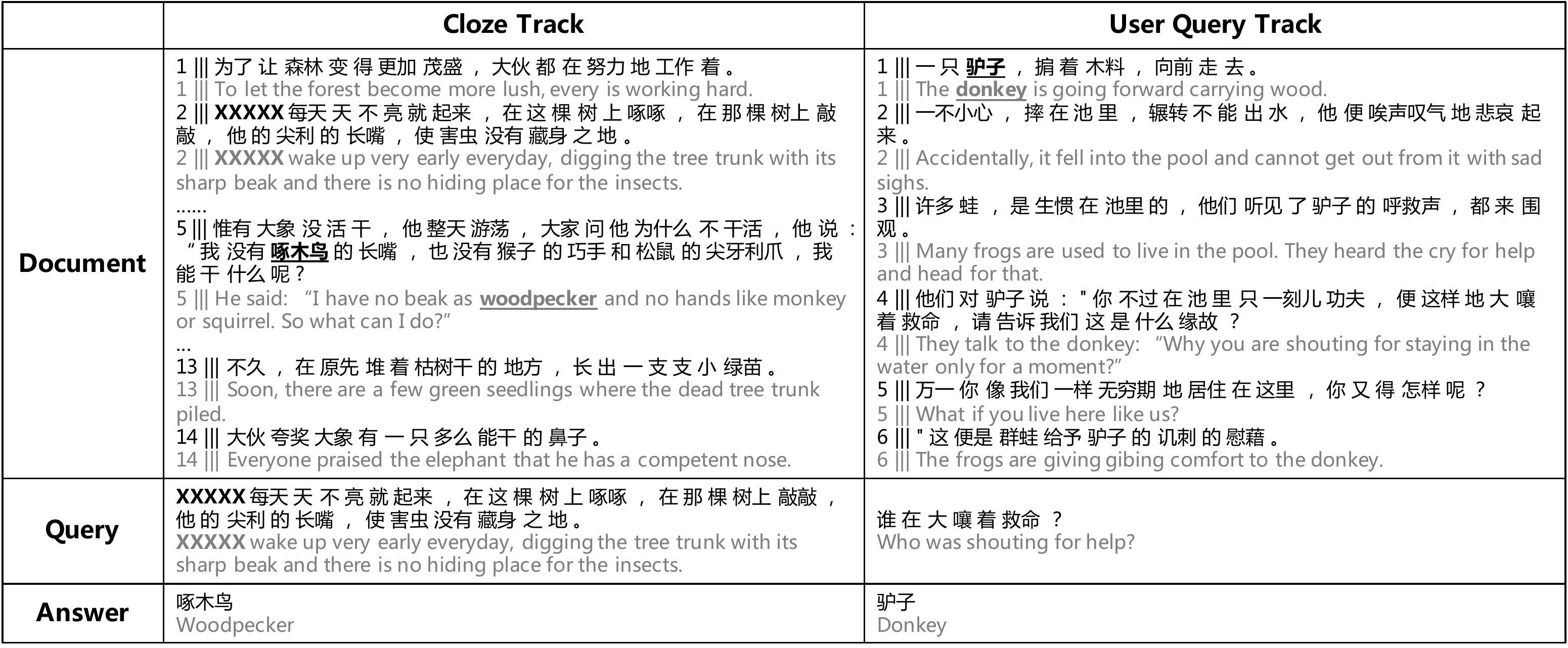}
  \caption{\label{data-examples} Examples of the proposed datasets (the English translation is in grey). The sentence ID is depicted at the beginning of each row. In the Cloze Track, ``XXXXX'' represents the missing word.  }
\end{figure*}

\section{The Proposed Dataset}\label{dataset}
In this section, we will briefly introduce the evaluation tracks and then the generation method of our dataset will be illustrated in detail. 

%%%%%%%%%%%%%%%%%%%%%%%
\subsection{The 1st Evaluation on Chinese Machine Reading Comprehension (CMRC-2017)}\label{cmrc2017}
The proposed dataset is typically used for the 1st Evaluation on Chinese Machine Reading Comprehension (CMRC-2017)\footnote{CMRC 2017 Official Website: {\url{http://www.hfl-tek.com/cmrc2017/index.html}}.}, which aims to provide a communication platform to the Chinese communities in the related fields. 
In this evaluation, we provide two tracks. We provide a shared training data for both tracks and separated evaluation data.

\begin{itemize}
	\item Cloze Track: In this track, the participants are required to use the large-scale training data to train their cloze system and evaluate on the cloze evaluation track, where training and test set are exactly the same type.
	\item User Query Track: This track is designed for using transfer learning or domain adaptation to minimize the gap between cloze training data and user query evaluation data, i.e. training and testing is fairly different.
\end{itemize}

Following Rajpurkar et al. \shortcite{rajpurkar-etal-2016}, we preserve the test set only visible to ourselves and require the participants submit their system in order to provide a fair comparison among participants and avoid tuning performance on the test set. The examples of Cloze and User Query Track are given in Figure \ref{data-examples}.

\subsection{Definition of Cloze Task}
The cloze-style reading comprehension can be described as a triple $\langle \mathcal D, \mathcal Q, \mathcal A \rangle$, where $\mathcal D$ represents {\bf D}ocument, $\mathcal Q$ represents {\bf Q}uery and the $\mathcal A$ represents {\bf A}nswer. 
There is a restriction that the answer should be a single word and should appear in the document, which was also adopted in \cite{hill-etal-2015,cui-etal-2016}.
In our dataset, we mainly focus on answering common nouns and named entities which require further comprehension of the document.

\subsection{Automatic Generation}
Following Cui et al. \cite{cui-etal-2016}, we also use similar way to generate our training data automatically.
Firstly we roughly collected 20,000 passages from children's reading materials which were crawled in-house.
Briefly, we choose an answer word in the document and treat the sentence containing answer word as the query, where the answer is replaced by a placeholder ``XXXXX''.
The detailed procedures can be illustrated as follows.
\begin{itemize}
	\item {\bf Pre-processing}: For each sentence in the document, we do word segmentation, POS tagging and dependency parsing using LTP toolkit \cite{che2010ltp}.
	\item {\bf Dependency Extraction}: Extract following dependencies: COO, SBV, VOB, HED, FOB, IOB, POB\footnote{Full descriptions of abbreviations can be found at {\url{http://www.ltp-cloud.com/intro/en/\#dp_how}}.}, and only preserve the parts that have dependencies. 
	\item {\bf Further Filtering}: Only preserve SBV, VOB and restrict the related words not to be pronouns and verbs.
	\item {\bf Frequency Restriction}: After calculating word frequencies, only word frequency that greater than 2 is valid for generating question.
	\item {\bf Question Restriction}: Only five questions can be extracted within one passage.
\end{itemize}

\subsection{Human Annotation}
Apart from the automatically generated large-scale training data, we also provide human-annotated validation and test data to improve the estimation quality. 
The annotation procedure costs one month with 5 annotators and each question is cross-validated by another annotator.
The detailed procedure for each type of dataset can be illustrated as follows.

\subsubsection{Cloze-style Reading Comprehension}
For the validation and test set in cloze data, we first randomly choose 5,000 paragraphs each for automatically generating questions using the techniques mentioned above. Then we invite our resource team to manually select 2,000 questions based on the following rules. 

\begin{itemize}
	\item Whether the question is appropriate and correct
	\item Whether the question is hard for LMs to answer
	\item Only select one question for each paragraph
\end{itemize}

\subsubsection{User Query Reading Comprehension}
Unlike the cloze dataset, we have no automatic question generation procedure in this type. In the user query dataset, we asked our annotator to directly raise questions according to the passage, which is much difficult and time-consuming than just selecting automatically generated questions. We also assign 5,000 paragraphs for question annotations in both validation and test data. Following rules are applied in asking questions.
\begin{itemize}
	\item The paragraph should be read carefully and judged whether appropriate for asking questions
	\item No more than 5 questions for each passage
	\item The answer should be better in the type of nouns, named entities to be fully evaluated
	\item Too long or too short paragraphs should be skipped
\end{itemize}

%%%%%%%%%%%%%%%%%%%%%%%
\section{Experiments}\label{experiments}
In this section, we will give several baseline systems for evaluating our datasets as well as presenting several top-ranked systems in the competition.

        \begin{table*}[t]
        \begin{center}
        \begin{tabular}{c l c p{1.2cm}<{\centering} p{1.2cm}<{\centering} p{1.2cm}<{\centering} p{1.2cm}<{\centering}}
        \toprule
        & & \multicolumn{2}{c}{Single Model} & \multicolumn{2}{c}{Ensemble} \\
        Rank & System & Validation & Test & Validation & Test  \\
        \midrule
        - & Baseline - Random Guess & 1.65 & 1.67 & - & - \\
        - & Baseline - Top Frequency & 14.85 & 14.07 & - & - \\
        - & Baseline - AS Reader (default settings) & 76.05 & 77.67 & - & - \\
        - & Baseline - AoA Reader (default settings) & {\bf 77.20} & {\bf 78.63} & - & - \\
        \midrule
        1 & 6ESTATES  & 75.85 & 74.73 & {\bf 81.85} & {\bf 81.90} \\
        2 & Shanghai Jiao Tong Univeristy BCMI-NLP & 76.15 & {\bf 77.73} & 78.35 & 80.67 \\
        3 & XinkTech & 77.15 & 77.53 & 79.20 & 80.27 \\
        4 & East China Normal University (ECNU) & {\bf 77.95} & 77.40 & 79.45 & 79.70 \\
        5 & Ludong University  & 74.75 & 75.07 & 77.05 & 77.07 \\
        6 & Wuhan University (WHU) & 78.20 & 76.53 & - & - \\
        7 & Harbin Institute of Technology at Shenzhen (HITSZ) & 76.05 & 75.93 & - & - \\
        8 & HuoYan Technology & 73.55 & 75.77 & - & - \\
        9 & Wuhan University of Science and Technology (WUST) & 73.80 & 74.53 & - & - \\
        10 & Beijing Information Science and Technology University & 70.05 & 70.20 & - & - \\
        11 & Shanxi Univerisity (SXU-2) & 62.60 & 64.70 & 66.65 & 68.47 \\
        12 & Shenyang Aerospace University (SAU) & 63.15 & 65.80 & - & - \\
        13 & Shanxi University (SXU-1) & 64.85 & 64.67 & - & - \\
        14 & Zhengzhou Univerisity (ZZU) & 52.80 & 54.53 & - & - \\
        \bottomrule
        \end{tabular}
        \end{center}
        \caption{\label{result-cloze} Results on Cloze Track. The best baseline and participant systems are depicted in bold face.}
        \end{table*}

\subsection{Baseline Systems}
We set several baseline systems for testing basic performance of our datasets and provide meaningful comparisons to the participant systems.
In this paper, we provide four baseline systems, including two simple ones and two neural network models.
The details of the baseline systems are illustrated as follows.

\begin{itemize}
	\item	{\bf Random Guess}: In this baseline, we randomly choose one word in the document as the answer.
	\item {\bf Top Frequency}: We choose the most frequent word in the document as the answer.
	\item {\bf AS Reader}: We implemented Attention Sum Reader (AS Reader) \cite{kadlec-etal-2016} for modeling document and query and predicting the answer with the Pointer Network \cite{vinyals-etal-2015}, which is a popular framework for cloze-style reading comprehension. Apart from setting embedding and hidden layer size as 256, we did not change other hyper-parameters and experimental setups as used in Kadlec et al. \shortcite{kadlec-etal-2016}, nor we tuned the system for further improvements.
	\item {\bf AoA Reader}: We also implemented Attention-over-Attention Reader (AoA Reader) \cite{cui-acl2017-aoa} which is the state-of-the-art model for cloze-style reading comprehension. We follow hyper-parameter settings in AS Reader baseline without further tuning.
\end{itemize}

In the User Query Track, as there is a gap between training and validation, we follow \cite{liu-acl2017-zp} and regard this task as domain adaptation or transfer learning problem. The neural baselines are built by the following steps.
\begin{itemize}
	\item We first use the shared training data to build a general systems, and choose the best performing model (in terms of cloze validation set) as baseline.
	\item Use User Query validation data for further tuning the systems with 10-fold cross-validations.
	\item Increase dropout rate \cite{srivastava-etal-2014} to 0.5 for preventing over-fitting issue.
\end{itemize}

All baseline systems are chosen according to the performance of the validation set.

        \begin{table}[tbp]
        \begin{center}
        \begin{tabular}{l c p{1.5cm}<{\centering} p{1.5cm}<{\centering}}
        \toprule
        %& \multicolumn{2}{c}{User Query Track} \\
        System & Validation & Test \\
        \midrule
        Baseline - Random Guess & 1.50 & 1.47 \\
        Baseline - Top Frequency & 10.65 & 8.73 \\
        Baseline - AS Reader & - & 49.03 \\
        Baseline - AoA Reader & - & {\bf 51.53} \\
        \midrule 
        ECNU (Ensemble) & 90.45 & {\bf 69.53} \\
        ECNU (single model) & 85.55 & 65.77 \\
        Shanxi University (Team-3) & 47.80 & 49.07 \\
        Zhengzhou University & 31.10 & 32.53 \\
        \bottomrule
        \end{tabular}
        \end{center}
        \caption{\label{result-userquery} Results on User Query Track. Due to the using of validation data, we did not report its performance.}
        \end{table}     

\subsection{Participant Systems}
The participant system results\footnote{Full CMRC 2017 Leaderboard: {\url{http://www.hfl-tek.com/cmrc2017/leaderboard.html}}.} are given in Table \ref{result-cloze} and \ref{result-userquery}. 

\subsubsection{Cloze Track}
As we can see that two neural baselines are competitive among participant systems and AoA Reader successfully outperform AS Reader and all participant systems in single model condition, which proves that it is a strong baseline system even without further fine-tuning procedure. Also, the best performing single model among participant systems failed to win in the ensemble condition, which suggest that choosing right ensemble method is essential in most of the competitions and should be carefully studied for further performance improvements.

\subsubsection{User Query Track}
Not surprisingly, we only received three participant systems in User Query Track, as it is much difficult than Cloze Track. As shown in Table \ref{result-userquery}, the test set performance is significantly lower than that of Cloze Track, due to the mismatch between training and test data. The baseline results give competitive performance among three participants, while failed to outperform the best single model by ECNU, which suggest that there is much room for tuning and using more complex methods for domain adaptation.

%%%%%%%%%%%%%%%%%%%%%%%
\section{Conclusion}\label{conclusion}
In this paper, we propose a new Chinese reading comprehension dataset for the 1st Evaluation on Chinese Machine Reading Comprehension (CMRC-2017), consisting large-scale automatically generated training set and human-annotated validation and test set. Many participants have verified their algorithms on this dataset and tested on the hidden test set for final evaluation. The experimental results show that the neural baselines are tough to beat and there is still much room for using complicated transfer learning method to better solve the User Query Task. We hope the release of the full dataset (including hidden test set) could help the participants have a better knowledge of their systems and encourage more researchers to do experiments on.

%%%%%%%%%%%%%%%%%%%%%%%
\section{Acknowledgements}
We would like to thank the anonymous reviewers for  their thorough reviewing and providing thoughtful comments to improve our paper. 
We thank the Sixteenth China National Conference on Computational Linguistics (CCL 2017) and Nanjing Normal University for providing space for evaluation workshop.
Also we want to thank our resource team for annotating and verifying evaluation data.
This work was supported by the National 863 Leading Technology Research Project via grant 2015AA015409.

\section{Bibliographical References}
\bibliographystyle{lrec}
\bibliography{xample}

\end{document}